\documentclass{article}

\usepackage{microtype}
\usepackage{graphicx}
\usepackage{booktabs} 

\usepackage{hyperref}

\usepackage{url}

\usepackage{xspace}
\usepackage{amsfonts}
\usepackage{nicefrac}
\usepackage{microtype}
\usepackage{amsmath}
\usepackage{subfig}
\usepackage{bm}
\usepackage{algorithm}
\usepackage{algpseudocode}
\usepackage{wrapfig}
\usepackage{xcolor}

\usepackage{dsfont}

\newcommand{\J}{\mathcal{J}}

\newcommand{\E}{\mathbb{E}}
\newcommand{\R}{\mathbb{R}}
\newcommand{\Loss}{\mathcal{L}}
\newcommand{\Converter}{\Psi}

\DeclareMathOperator*{\argmax}{arg\,max}

\newcommand{\OurMethod}{$L^{3}P$\xspace}

\usepackage[accepted]{icml2021}

\icmltitlerunning{World Model as a Graph: Learning Latent Landmarks for Planning}

\begin{document}

\twocolumn[
\icmltitle{World Model as a Graph: \\Learning Latent Landmarks for Planning}



\icmlsetsymbol{equal}{*}

\begin{icmlauthorlist}
\icmlauthor{Lunjun Zhang}{toronto,vector}
\icmlauthor{Ge Yang}{goo}
\icmlauthor{Bradly Stadie}{ed}
\end{icmlauthorlist}

\icmlaffiliation{toronto}{University of Toronto}
\icmlaffiliation{vector}{Vector Institute}
\icmlaffiliation{goo}{MIT}
\icmlaffiliation{ed}{Toyota Technological Institute at Chicago}

\icmlcorrespondingauthor{Lunjun Zhang}{lunjun@cs.toronto.edu}

\icmlkeywords{Machine Learning, ICML}

\vskip 0.3in
]



\printAffiliationsAndNotice{}  

\begin{abstract}
Planning, the ability to analyze the structure of a problem in the large and decompose it into interrelated subproblems, is a hallmark of human intelligence. 
While deep reinforcement learning (RL) has shown great promise for solving relatively straightforward control tasks, it remains an open problem how to best incorporate planning into existing deep RL paradigms to handle increasingly complex environments.
One prominent framework, Model-Based RL, learns a world model and plans using step-by-step virtual rollouts. This type of world model quickly diverges from reality when the planning horizon increases, thus struggling at long-horizon planning. How can we learn world models that endow agents with the ability to do temporally extended reasoning? In this work, we propose to learn graph-structured world models composed of sparse, multi-step transitions. We devise a novel algorithm to learn latent landmarks that are scattered (in terms of reachability) across the goal space as the nodes on the graph. In this same graph, the edges are the reachability estimates distilled from Q-functions. On a variety of high-dimensional continuous control tasks ranging from robotic manipulation to navigation, we demonstrate that our method, named \OurMethod, significantly outperforms prior work, and is oftentimes the only method capable of leveraging both the robustness of model-free RL and generalization of graph-search algorithms.
We believe our work is an important step towards scalable planning in reinforcement learning. 

\end{abstract}

\section{Introduction}

An intelligent agent should be able to solve difficult problems by breaking them down into sequences of simpler problems. Classically, planning algorithms have been the tool of choice for endowing AI agents with the ability to reason over complex long-horizon problems \citep{doran1966experiments, hart1968formal}. Recent years have seen an uptick in monographs examining the intersection of classical planning techniques -- which excel at temporal abstraction -- with deep reinforcement learning (RL) algorithms -- which excel at state abstraction. Perhaps the ripest fruit born of this relationship is the AlphaGo algorithm, wherein a model free policy is combined with a MCTS \citep{mcts} planning algorithm to achieve superhuman performance on the game of Go \citep{silver2016mastering}. 

In the field of robotics, progress on combining planning and reinforcement learning has been somewhat less rapid, although still resolute. Indeed, the laws of physics in the real world are vastly more complex than the simple rules of Go. Unlike board games such as chess and Go, which have deterministic and known dynamics and discrete action space, robots have to deal with a probabilistic and unpredictable world. Moreover, the action space for robotics is often continuous. As a result of these difficulties, planning in robotics presents a much harder problem. 
One general class of methods \citep{dyna} seeks to combine model-based planning and deep RL. These methods can be thought of as an extension of model-predictive control (MPC) algorithms, with the key difference being that the agent is trained over hypothetical experience in addition to the actually collected experience. The primary shortcoming of this class of methods is that, like MCTS in AlphaGo, they resort to planning with action sequences -- forcing the robot to plan for each action at every hundred milliseconds. 
Planning on the level of action sequences is fundamentally bottlenecked by the accuracy of the learned dynamics model and the horizon of a task, as the learned world model quickly diverges over a long horizon. This limitation shows that world models in the traditional Model-based RL (MBRL) setting often fail to deliver the promise of planning. 

Another general class of methods, Hierarchical RL (HRL),  introduces a higher-level learner to address the problem of planning \citep{feudal_hinton, vezhnevets2017feudal, hiro}. In this case, a goal-based RL agent serves as the worker, and a manager learns what sequences of goals it must set for the worker to achieve a complex task. 
While this is apparently a sound solution to the problem of planning, hierarchical learners neither explicitly learn a higher-level model of the world nor take advantage of the graph structure inherent to the problem of search. 

\begin{figure}
    \centering
    \includegraphics[width=0.43\textwidth]{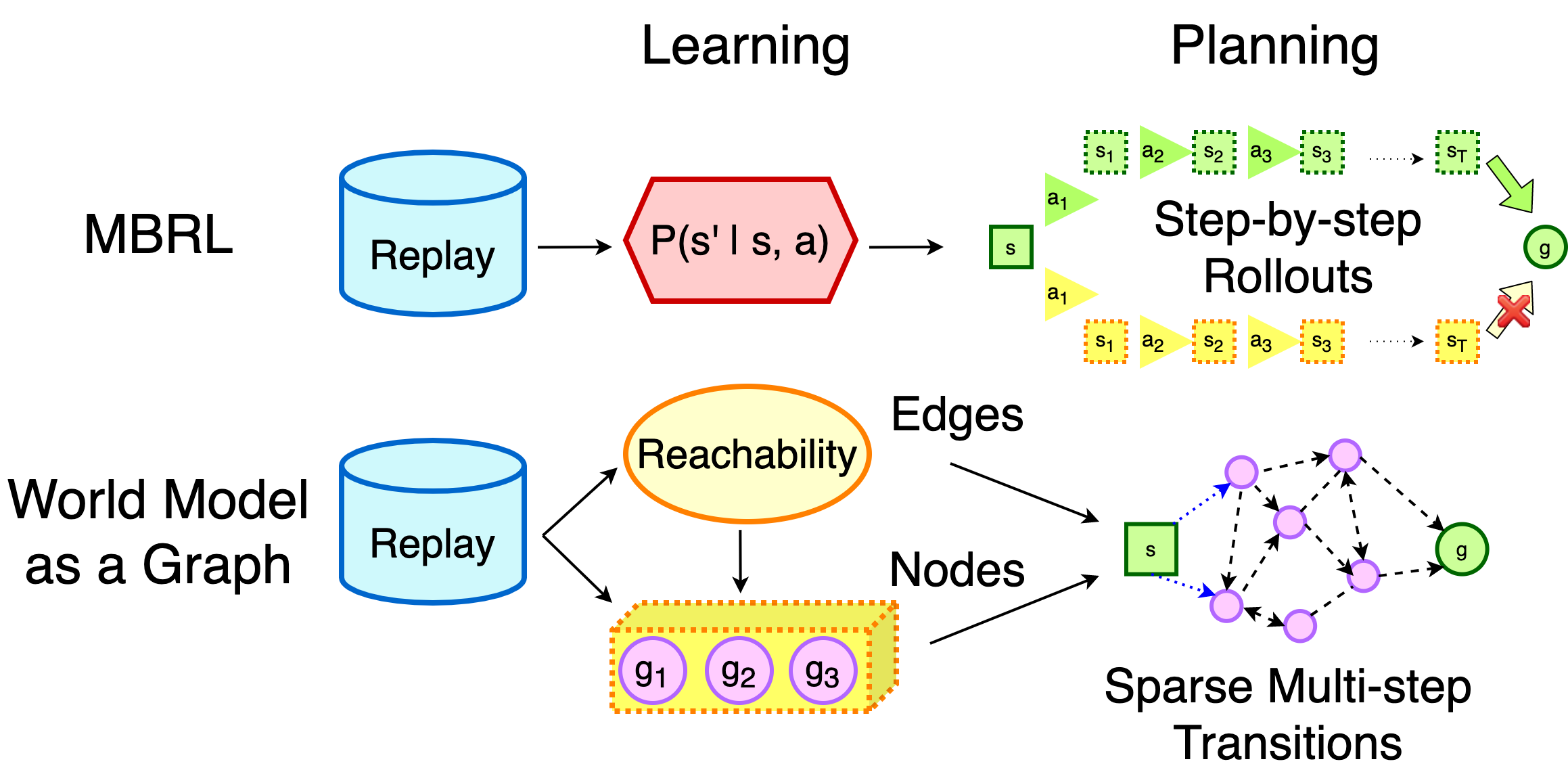}
    \caption{MBRL versus \OurMethod (World Model as a Graph). MBRL does step-by-step virtual rollouts with the world model and quickly diverges from reality when the planning horizon increases. \OurMethod models the world as a graph of sparse multi-step transitions, where the nodes are learned latent landmarks and the edges are reachability estimates. \OurMethod succeeds at temporally extended reasoning. \textbf{Code} for \OurMethod is available at: \url{https://github.com/LunjunZhang/world-model-as-a-graph}.}
    \label{fig:overview}
    \vspace{-4mm}
\end{figure}

To better combine classical planning and reinforcement learning, we propose to learn graph-structured world models composed of sparse multi-step transitions. To model the world as a graph, we borrow a concept from the navigation literature -- the idea of landmarks \citep{wang2008robot}. Landmarks are essentially states that an agent can navigate between in order to complete tasks. However, rather than simply using previously seen states as landmarks, as is traditionally done, we will instead develop a novel algorithm to learn the landmarks used for planning. Our key insight is that by mapping previously achieved goals into a latent space that captures the temporal distance between goals, we can perform clustering in the latent space to group together goals that are easily reachable from one another. Subsequently, we can then decode the latent centroids to obtain a set of goals scattered (in terms of reachability) across the goal space. Since our learned landmarks are obtained from latent clustering, we call them \textit{latent landmarks}. The chief algorithmic contribution of this paper is a new method for planning over learned latent landmarks for high-dimensional continuous control domains, which we name Learning Latent Landmarks for Planning (\OurMethod). 


The idea of reducing planning in RL to a graph search problem has enjoyed some attention recently \citep{sptm, sorb, mss, Liu2019-mh,  yang2020plan2vec, sgm}. A key difference between those works and \OurMethod is that our use of \textit{learned} latent landmarks allows us to substantially reduce the size of the search space. What’s more, we make improvements to the graph search module and the online planning algorithm to improve the robustness and sample efficiency of our method. As a result of those decisions, our algorithm is able to achieve superior performance on a variety of robotics domains involving both navigation and manipulation. In addition to the results presented in Section \ref{sec:experiments}, \textbf{videos} of our algorithm’s performance and a more detailed analysis of the sub-tasks discovered by the latent landmarks can be found at: \url{https://sites.google.com/view/latent-landmarks/}. 

\begin{figure*}[h]
    \centering
    \includegraphics[width=0.75\textwidth]{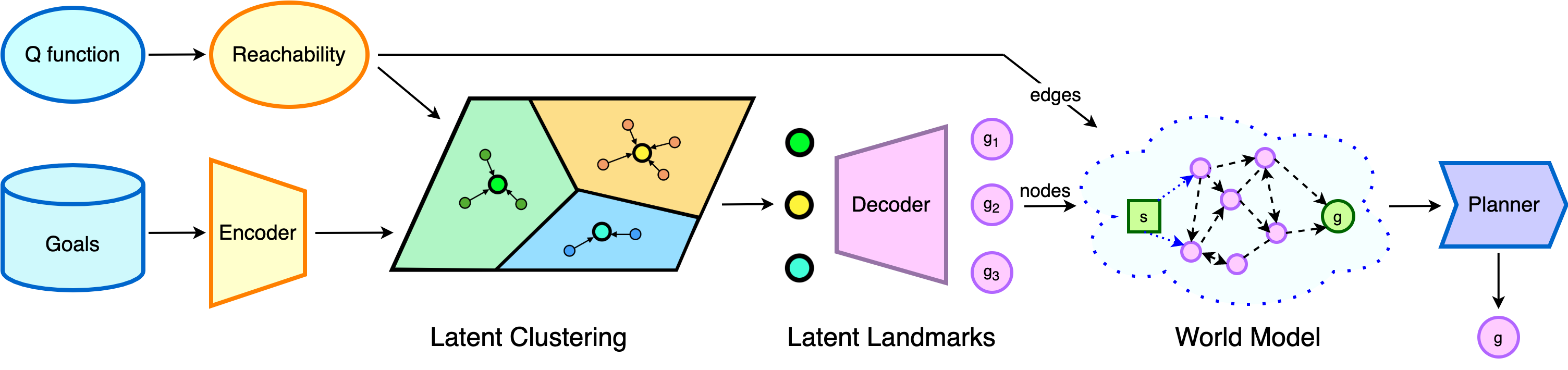}
    \caption{An overview of \OurMethod, which learns a small number of latent landmarks for planning. The main components of our method are: learning reachability estimates (via Q-learning and regression), learning a latent space (via an auto-encoder with reachability constraints), learning latent landmarks (via clustering in the latent space), graph search on the world model and online planning.}
    \label{fig:overview}
    \vspace{-3mm}
\end{figure*}

\section{Related Works}

The problem of learning landmarks to aid in robotics problems has a long and rich history \citep{gillner1998navigation, wang2002human, wang2008robot}. Prior art has been deeply rooted in the classical planning literature. For example, traditional methods would utilize \cite{dijkstra1959note} to plan over generated waypoints, SLAM \citep{durrant2006simultaneous} to simultaneously integrate mapping, or the RRT algorithm \citep{lavalle1998rapidly} for explicit path planning. The A* algorithm \citep{hart1968formal} further improved the computational efficiency of Dijkstra. Those types of methods often heavily rely on a hand-crafted configuration space that provides prior knowledge. 

Planning is intimately related to model-based RL (MBRL), as the core ideas underlying learned models and planners can enjoy considerable overlap. Perhaps the most clear instance of this overlap is Model Predictive Control (MPC), and the related Dyna algorithm \citep{dyna}. When combined with modern techniques \citep{me_trpo, slbo, mb_mf, ha2018recurrent, latent_dynamics, exploring_mbrl, mbpo}, MBRL is able to achieve some level of success. \cite{state_tabulation} and \cite{hafner2020mastering} also learn a discrete latent representation of the environment in the MBRL framework. As discussed in the introduction, planning on action sequences will fundamentally struggle to scale in robotics. 

Our method makes extensive use of a parametric goal-based RL agent to accomplish low-level navigation between states. This area has seen rapid progress recently, largely stemming from the success of Hindsight Experience Replay (HER) \citep{her}. Several improvements to HER augment the goal relabeling and sampling strategies to boost performance \citep{rig, tdm, skew-fit, mep, mega}. There have also been attempts at incorporating search as inductive biases within the value function \citep{Silver2016predictron, tamar2016VIN, Farquhar2017treeQN, Racaniere2017I2A, Lee2018gppn,  Srinivas2018UPN}. The focus of this line of work is to improve the low-level policy and is thus orthogonal to our work. 

Recent work in Hierarchical RL (HRL) builds upon goal-based RL by learning a high-level parametric manager that feeds goals to the low-level goal-based agent \citep{feudal_hinton, vezhnevets2017feudal, hiro}. This can be viewed as a parametric alternative to classical planning, as discussed in the introduction. Recently, \cite{subgoal-trees, long_plan} have derived HRL methods that are intimately tied to tree search algorithms. These papers are further connected to a recent trend in the literature wherein classical search methods are combined with parametric control \citep{sptm, sorb, mss, Liu2019-mh,  yang2020plan2vec, sgm}. Several of these articles will be discussed throughout this paper. LEAP \citep{leap} also considers the problem of proposing sub-goals for a goal-conditioned agent: it uses a VAE \citep{vae} and does CEM on the prior distribution to form the landmarks. Our method constrains the latent space with temporal reachability between goals, a concept previously explored in \cite{reachability}, and uses latent clustering and graph search rather than sampling-based methods to learn and propose sub-goals. 

\section{Background}

We consider the problem of Multi-Goal RL under a Markov Decision Process (MDP) parameterized by $(S, A, \mathds P, G, \Converter, R, \rho_0)$. $S$ and \(A\) are the state and action space.  The probability distribution of the initial states is given by $\rho_0(s)$, and $\mathds P(s'\vert s, a)$ is the transition probability. $ \Converter: S \mapsto G$ is a mapping from the state space to the goal space, which assumes that \textit{every} state \(s\) can be mapped to a corresponding \textit{achieved} goal \(g\). The reward function \(R\) can be defined as $R(s,a,s', g) = -\mathds{1} \{\Converter(s') \neq g \}$. We further assume that each episode has a fixed horizon $T$. 

A multi-goal policy is a probability distribution $\pi: S \times G \times A \rightarrow \R^{+}$, which gives rise to trajectory samples of the form $\tau = \{s_{0}, a_{0}, g, s_{1}, \cdots s_{T}\}$. The purpose of the policy $\pi$ is to learn how to reach the goals drawn from the goal distribution $p_{g}$. With a discount factor $\gamma \in (0, 1)$, it maximizes $\J(\pi) = \E_{g \sim p_{g}, \tau \sim \pi(g)}[\sum_{t=0}^{T-1} \gamma^{t} \cdot R(s_{t}, a_{t}, s_{t+1}, g) ]$. Q-learning provides a sample-efficient way to optimize the above objective by utilizing off-policy data stored in a replay buffer $B$. $Q(s,a,g)$ estimates the reward-to-go under the current policy $\pi$ conditioned upon the given goal. An additional technique, called Hindsight Experience Replay, or HER \citep{her}, uses hindsight relabelling to drastically speed up training. This relabeling crucially relies upon the mapping $\Converter: S \mapsto G$ in the multi-goal MDP setting. We can write the the joint objective of multi-goal Q-learning with HER as minimizing (with $Q$ being the online network and $\widehat{Q}$ being the target network):
\begin{align}
    \Big(Q(s_{t}, a_{t}, g) - \Big(R(s_{t+1}, g) + \gamma \cdot \widehat{Q}(s_{t+1}, a', g)\Big) \Big)^{2} \label{dqn-loss}
\end{align}
where $\tau \sim B, t\sim \{0\cdots T-1\}, (s_{t},a_{t},s_{t+1}) \sim \tau, k \sim \{t+1 \cdots T\} , g = \Converter(s_{k}) , a' \sim \pi(\cdot \mid s_{t+1}, g) $. 

\section{The \OurMethod \ Algorithm}
\vspace{-2mm}

Our overall objective in this section is to derive an algorithm that learns a small number of landmarks scattered across goal space in terms of reachability and use those learned landmarks for planning. There are three chief difficulties we must overcome when considering such an algorithm. First, how can we group together goals that are easily reachable from one another? The answer is to embed goals into a latent space, where the latent representation captures some notion of temporal distance between goals -- in the sense that goals that would take many timesteps to navigate between are further apart in latent space. Second, we need to find a way to learn a sparse set of landmarks used for planning. Our method performs clustering on the constrained latent space, and decodes the learned centroids as the landmarks we seek. Finally, we need to develop a non-parametric planning algorithm responsible for selecting sequences of landmarks the agent must traverse to accomplish its high-level goal. The proposed online planning algorithm is simple, scalable, and robust. 

\subsection{Learning a Latent Space}
\label{sec:latent-space}
Let us consider the following question: ``How should we go about learning a latent space of  goals where the metric reflects reachability?'' Suppose we have an auto-encoder (AE) in the agent's goal space, with deterministic encoder $f_E$ and decoder $f_D$. As usual, the reconstruction loss is given by $\Loss_{rec}(g) = \Big\lVert f_{D} \big(f_{E}(g)\big) - g \Big\rVert_{2}^{2}$. We want to make sure that the distance between two latent codes would roughly correspond to the number of steps it would take the policy to go from one goal to another. Concretely, for any pair of goals $(g_{1}, g_{2})$, we optimize the following loss $\Loss_{latent}(g_{1}, g_{2})$:
\begin{align}
    \label{eq:ae-loss}
    \Big( \big\lVert f_{E}(g_{1}) - f_{E}(g_{2}) \big\rVert_{2}^{2} - \dfrac{1}{2} \big(V(g_{1}, g_{2}) + V(g_{2}, g_{1})\big) \Big)^{2}
\end{align}
Where $V: G \times G \rightarrow \R^+$ is a mapping that estimates how many steps it would take the policy $\pi$ to go from one goal to another goal on average. By adding this constraint and solving a joint optimization $\Loss_{rec} + \lambda \cdot \Loss_{latent}$, the encoding-decoding mapping can no longer be arbitrary, giving more structure to the latent space. Goals that are close by in terms of reachability will be naturally clustered in the latent space, and interpolations between latent codes will lead to meaningful results. 

Of course, the constraint in Equation \ref{eq:ae-loss} is quite meaningless if we do not have a way to estimate the mapping $V$. We will proceed towards this objective by noting the following interesting connection between multi-goal Q-functions and reachability. In the multi-goal RL framework considered in the background section, the reward is binary in nature. The agent receives a reward of $-1$ until it reaches the goal, and then $0$ when it reaches the desired goal. In this setting, the Q-function is implicitly estimating \textit{the number of steps} it takes to reach the goal $g$ from the current state $s$ \textit{after} the action $a$ is taken. Denote this quantity as $D(s,a,g)$, the Q-function can be re-written as:
\begin{equation}
\begin{aligned}
    Q(s,a,g) &= \sum_{t=0}^{D(s,a,g) - 1} \gamma^{t} \cdot (-1) + \sum_{t=D(s,a,g)}^{T-1} \gamma^{t} \cdot 0 \\
    &= - \dfrac{1 - \gamma^{D(s,a,g)}}{1 - \gamma} \label{dis-function}
\end{aligned}
\end{equation}
Choosing to parameterize Q-functions in this way disentangles the effect of $\gamma$ on multi-goal Q-learning. It also provides us with access the direct distance estimation function $D(s,a,g)$. We note that this \textit{distance} is not a mathematical distance in the sense of a metric. Instead, we use the word \textit{distance} to refer to the number of steps the policy $\pi$ needs to take in the environment. 

Given our tractable estimate of $D$, it is now a straightforward matter to estimate the desired quantity $V$, which approximates how many steps it takes the policy to transition between goals. To get the desired estimate, we regress $V$ towards $D$ by minimizing 
\begin{align}
    \min_{V} \Bigg( D\big(s_{t}, a_{t}, \Converter(s_{k})\big) - V\big(\Converter(s_{t+1}), \Converter(s_{k})\big) \Bigg)^{2}
\end{align}
with $\tau \sim B, t\sim \{0\cdots T-1\}, (s_{t},a_{t},s_{t+1}) \sim \tau, k \sim \{t+1 \cdots T\}$, and $\Converter$ being given by the environment to map the states to the goal space. One crucial detail is the use of $\Converter(s_{t+1})$ rather than $\Converter(s_{t})$ in the inputs to $V$. This is due to the fact that $D: S \times A \times G \rightarrow \R$ outputs the number of steps to go \textit{after} an action is taken, when the state has transitioned into $s_{t+1}$. The objective above provides an unbiased estimate of the average number of steps between two goals. 

The estimates $D$ and $V$ will prove useful beyond helping to optimize the auto-encoder in Equation \ref{eq:ae-loss}. They will prove essential in weighting and planning over latent landmark nodes in Section 4.3.

\subsection{Learning Latent Landmarks}
\vspace{-2mm}

Planning on a graph can be expensive, as the number of edges can grow quadratically with the number of nodes. To battle this issue in scalability, we use the constrained latent space to learn a sparse set of landmarks. A landmark can be thought of as a waypoint that the agent can pass through enroute to achieve a desired goal. Ideally, \textit{goals that are easily reachable from one another should be grouped to form one single landmark}. Since our latent representation captures the temporal reachability between goals, this can be achieved by doing clustering in the latent space. The cluster centroids, when decoded from the decoder, will be precisely the latent landmarks we are seeking. 

Clustering proceeds as follows. For $N$ clusters to be learned, we define a mixture of Gaussians in the latent space with $N$ trainable latent centroids, $\{\textbf{c}_{1} \cdots \textbf{c}_{N}\}$, and a shared trainable variance vector $\boldsymbol{\sigma}$. We maximize the evidence lower bound (ELBO) with a uniform prior $p(\textbf{c})$:
\begin{equation}
\begin{aligned}
    &\log p\Big(z = f_{E}(g) \Big) \\
    & \geq \E_{q(\textbf{c} \mid z)}\Big[\log p(z \mid \textbf{c})\Big] - D_{KL}\Big(q(\textbf{c} \mid z) \parallel p(\textbf{c})\Big) \label{elbo_eq}
\end{aligned} 
\end{equation}
Ideally, we would like each batch of data given to the latent clustering model to be representative of the whole replay buffer, such that the centroids will quickly learn to scatter out. To this end, we propose to use the Greedy Latent Sparsification (GLS) algorithm (see the Appendix) on each batch of data sampled from the replay before taking a gradient step with the batch. GLS is inspired by kmeans++ \citep{kmeans_plus_plus}, with several key differences: this sparsification process is used for both training and initialization,  it uses a neural metric for determining the distance between data points, and that it is compatible with mini-batch-style gradient-based training. 

\label{section-learning-latent-landmarks}

\subsection{Planning with Latent Landmarks}

\label{section-planning-with-latent-landmarks}
Having derived a latent encoding algorithm and an algorithm for learning latent landmarks, we at last turn our attention to search and planning. \OurMethod is agnostic to the graph search algorithm being used. In practice, we use a variant of the Floyd algorithm, where our relaxation operations use a soft max rather than hard max for better stability (see the Appendix for more details). To construct a weight matrix that provides raw distance estimates between latent landmarks in the first place, we begin by decoding the learned centroids in the latent space into the nodes in the graph $\{f_{D}(\textbf{c}_{1}) \cdots f_{D}(\textbf{c}_{N})\}$. To build the graph, we add two edges directed in reverse orders for every pair of latent landmarks. For instance, for an edge going from $f_{D}(\textbf{c}_{i})$ to $f_{D}(\textbf{c}_{j})$, the weight on that edge is $w_{i,j} = -V(f_{D}(\textbf{c}_{i}), f_{D}(\textbf{c}_{j}))$. Notice that the distances are negated. At the start of an episode, the agent receives a goal $g$, and we construct matrix $W$:
\begin{align}
    W = \begin{pmatrix} 
        0 & \dots & w_{1,N} & -V(f_{D}(\textbf{c}_{1}), g)\\ 
        \vdots & \ddots & \vdots & \vdots\\
        w_{N, 1} & \dots & 0 & -V(f_{D}(\textbf{c}_{N}), g) \\
        -\infty & \dots & -\infty & 0
    \end{pmatrix} \label{distance_matrix}
\end{align}
\begin{minipage}{0.48\textwidth}
\begin{algorithm}[H]
    \caption{Online Planning in \OurMethod}
    \textbf{Given}: Environment \texttt{env}, initial state $s$, goal $g$.
    \begin{algorithmic}[1]
        \State \texttt{Cnt} = $0$. \texttt{SubG} = \texttt{None}. 
        \State Solve for $\boldsymbol{d}_{\boldsymbol{c} \rightarrow g}$ with \textcolor{red}{\textbf{graph search}} using $W$. 
        \For{$t$ = $1$ to $T$} \Comment{One episode}
            \If {\texttt{Cnt} $\geq 1.0$}
                \State \texttt{Cnt} $=\texttt{Cnt}- 1$
            \Else \Comment{\textcolor{blue}{We do \textbf{not} re-plan at  every step}}
                \State Calculate $\boldsymbol{d}_{s\rightarrow \boldsymbol{c}}$.
                \State $\boldsymbol{d}$ $\leftarrow$ $\boldsymbol{d}_{s\rightarrow \boldsymbol{c}} + \boldsymbol{d}_{\boldsymbol{c} \rightarrow g}$
                \State \Comment{\textcolor{blue}{Remove the \textbf{immediate} previous landmark}}
                \If {\texttt{SubG} $\neq$ \texttt{None}}
                    \State $\boldsymbol{d}[\texttt{SubG}] \leftarrow -\infty$
                \EndIf 
                \State \texttt{SubG}, \texttt{Cnt} $\leftarrow \argmax \boldsymbol{d}$, $-\max \boldsymbol{d}$ 
            \EndIf
            \State $a \sim \pi(s, \texttt{SubG})$; $s \leftarrow$ \texttt{env.step(a)}.
        \EndFor
    \end{algorithmic} \label{online_planning}
\end{algorithm} 
\end{minipage}\hfill

\begin{figure}
    \centering
    \includegraphics[width=0.4\textwidth]{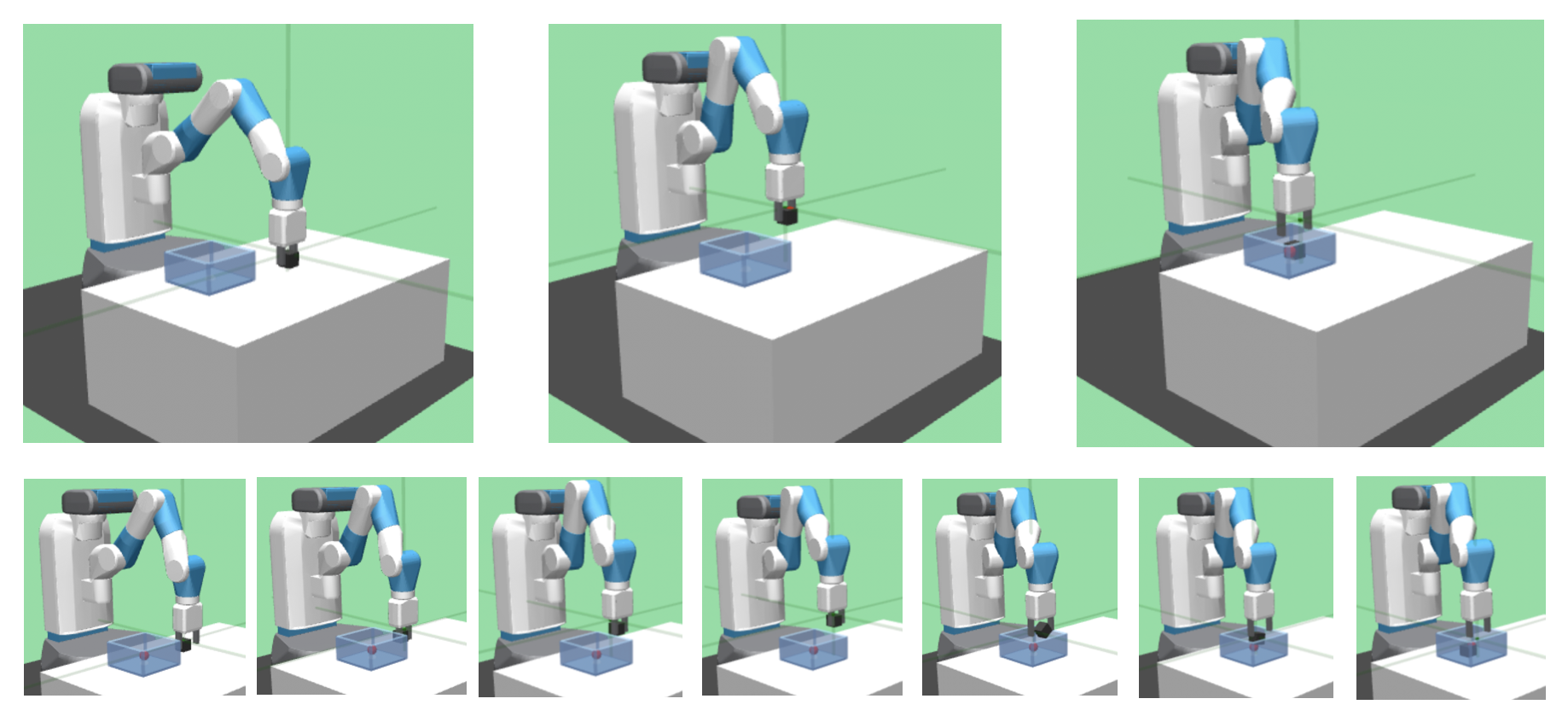}
    \caption{We consider two environments involving a fetch robot, a block, and a box. In Box-Distractor-PickAndPlace, the fetch must learn to pick and place the block while avoiding collision with the box. In Place-Inside-Box, the fetch must pick the block and place it inside the box. We visualize the fetch states corresponding to learned landmarks in the second row of images. } 
    \label{fig:place-task}
\end{figure}

For online planning, when the agent receives a goal at the start of an episode, we use graph search to solve for $\boldsymbol{d}_{\boldsymbol{c}\rightarrow g}$ (which is fixed throughout an episode). For an observation state $s$, the algorithm calculates $\boldsymbol{d}_{s\rightarrow \boldsymbol{c}}$:
\begin{align}
    \boldsymbol{d}_{s\rightarrow \boldsymbol{c}} &= \begin{pmatrix} 
        - D\big(s, \pi(s, f_{D}(\textbf{c}_{1})), f_{D}(\textbf{c}_{1})\big) \\ 
        \vdots \\
        - D\big(s, \pi(s, f_{D}(\textbf{c}_{N})), f_{D}(\textbf{c}_{N})\big) \\
        - D\big(s, \pi(s, g), g\big)
    \end{pmatrix} \label{state_to_landmarks}
\end{align}
The chosen landmark is $\texttt{subgoal} \leftarrow \argmax (\boldsymbol{d}_{s\rightarrow \boldsymbol{c}} + \boldsymbol{d}_{\boldsymbol{c} \rightarrow g})$. To further provide temporal abstraction and robustness, the agent will be asked to consistently pursue \texttt{subgoal} for $K=-\boldsymbol{d}_{s\rightarrow \boldsymbol{c}}[\texttt{subgoal}]$ number of steps, which is \textit{how many steps it thinks it will need}. The proposed goal does \textbf{not} change during this period. In this way, \OurMethod makes sure that the agent does not re-plan at every step, and this mechanism for temporal abstraction is crucial to its robustness. This mechanism is detailed in Algorithm \ref{online_planning}.

After this $K$ many steps, the agent will decide on the next landmark to pursue by re-calculating $\boldsymbol{d}_{s\rightarrow \boldsymbol{c}}$, but the \textit{immediate} previous landmark will not be considered as a candidate landmark. The reason is that, if the agent has failed to reach a self-proposed landmark within the reachability limit it has set for itself, then the agent should try something new for the immediate next goal rather than stick to the immediate previous landmark for another round. We have found that this simple algorithm helps the agent avoid getting stuck and improves the overall robustness of the agent. 

In summary, we have derived an algorithm that learns a sparse set of latent landmarks scattered across goal space in terms of reachability, and uses those learned landmarks for robust planning. 

\section{Experiments and Evaluation}
\label{sec:experiments}
We investigate the impact of \OurMethod in a variety of robotic manipulation and navigation environments. These include standard benchmarks such as Fetch-PickAndPlace, and more difficult environments such as AntMaze-Hard and Place-Inside-Box that have been engineered to require test-time generalization. Videos of our algorithm in action are available at: \url{https://sites.google.com/view/latent-landmarks/}.

\begin{figure*}[h]
    \centering
    \includegraphics[width=0.9\textwidth]{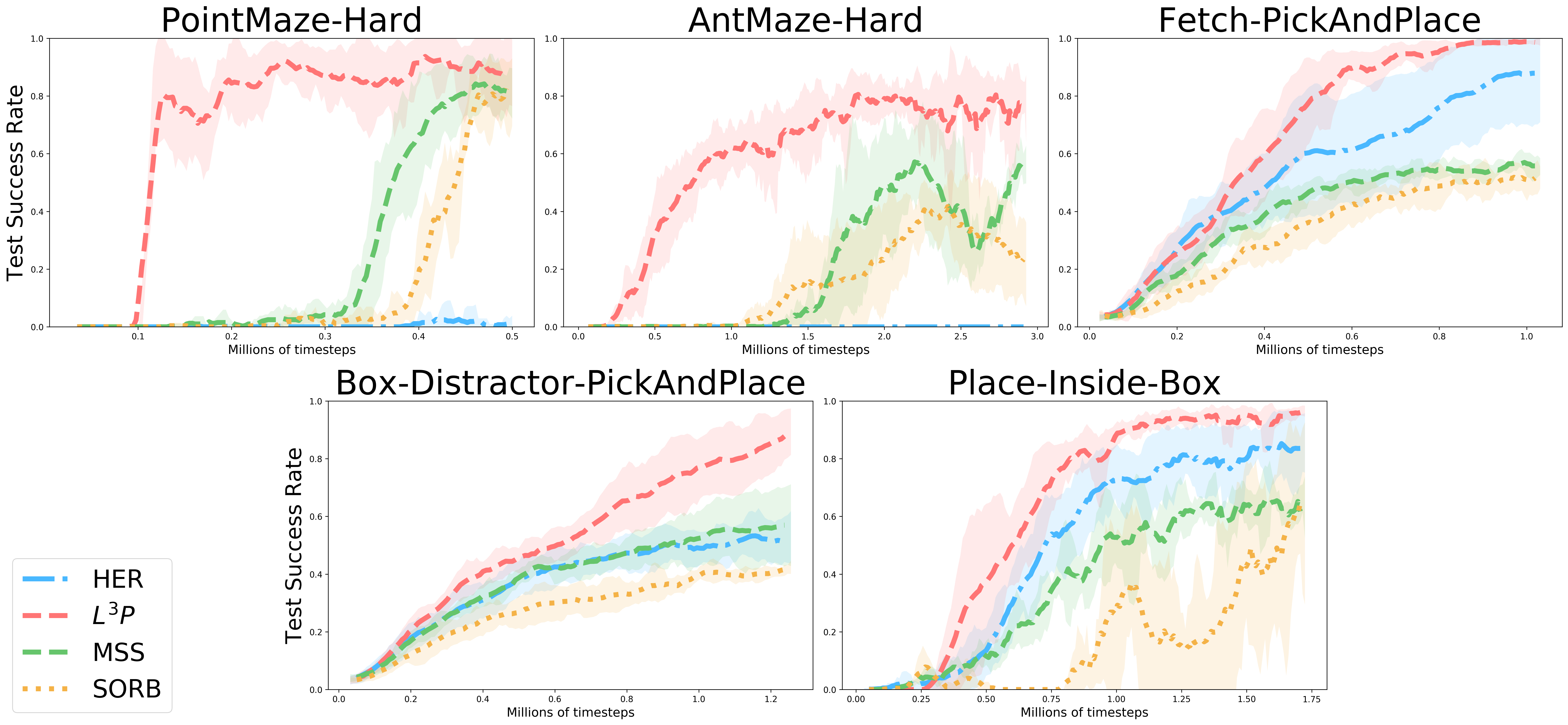}
    \label{fig:test_curves}
    \caption{Test time success rate vs. total number of timesteps, on a variety of challenging  robotic navigation and manipulation environments. \OurMethod demonstrates better sample efficiency, higher asymptotic performance, and in some cases, the ability to generalize to longer horizons. }
\end{figure*}

\begin{figure}
    \centering
    \includegraphics[width=0.45\textwidth]{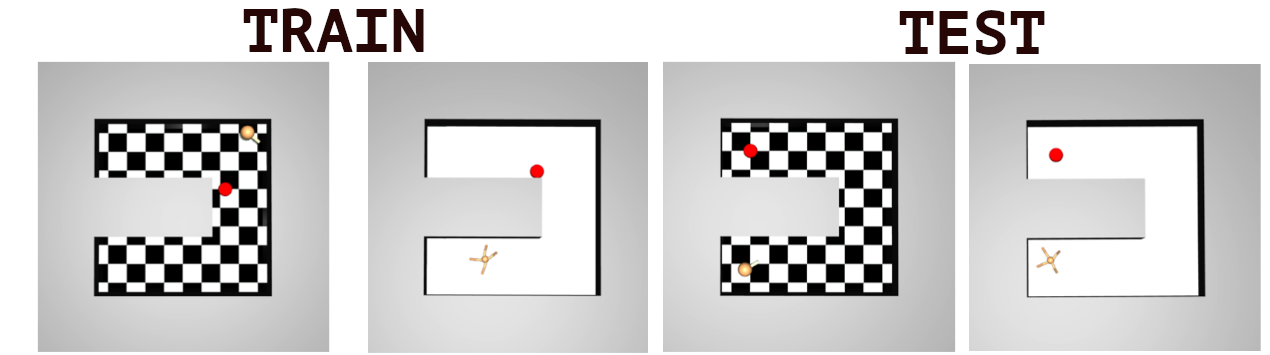}
    \caption{For both Point and Ant, during training, the initialization state distribution and the goal proposal distribution are \textit{uniform} around the maze. During test time, the agent is asked to traverse the longest path in the maze, which is not seen during training. Importantly, the map of the environment is not given to the agent at any given point; the agent has to learn the structure of the environment purely through interaction. The success rate during test is reported in Figure \ref{fig:test_curves}. This environment demonstrates \OurMethod's ability to generalize to longer horizon goals during test time.}
    \label{fig:mazes}
    \vspace{-3mm}
\end{figure}

\begin{figure}
    \centering
    \includegraphics[width=0.48\textwidth]{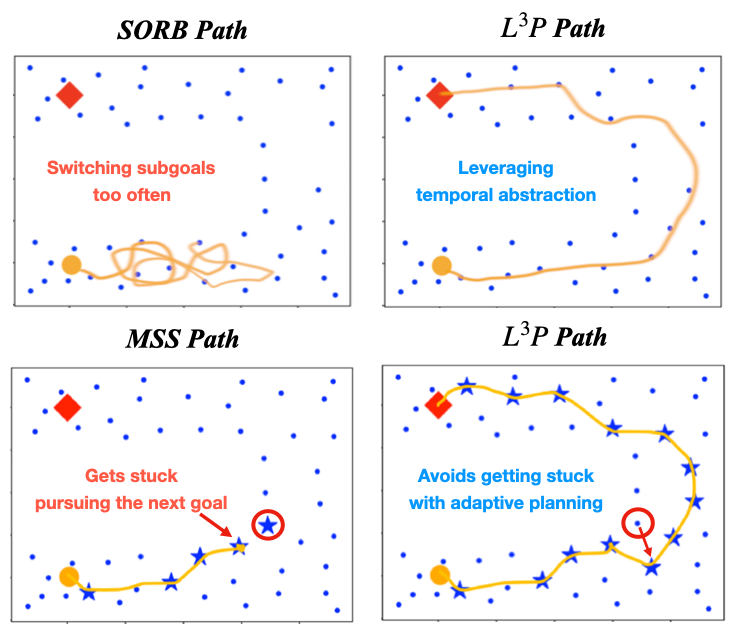}
    \caption{Visualizing the paths taken by SORB, MSS and \OurMethod on AntMaze at test time. The \textcolor{blue}{blue dots} in the backgrounds are the learned landmarks using \OurMethod. The \textcolor{orange}{orange dot} is the starting location of the Ant. The \textcolor{red}{red dot} is the final goal. The \textcolor{blue}{blue stars} indicate the landmarks chosen by the planning algorithms. As illustrated in the figure above, \OurMethod addresses two major failure modes of graph-based planning with RL. Firstly, graph-based methods tend to switch proposed subgoals too frequently and fall into a loop due to wormholes in distance estimates, whereas \OurMethod leverages temporal abstraction in both landmark learning and online planning to avoid this pitfall. Secondly, when the agent pursues a subgoal unsuccessfully (due to obstacles, etc), other methods tend to get stuck by continuing proposing the same subgoal, whereas \OurMethod can adapt to the encountered failure and propose different subgoals in the event of getting stuck. 
    }
    \label{fig:plan_visualize}
\end{figure}

\vspace{-2mm}

    

\subsection{Baselines}
\vspace{-2mm}
We compare our method with a variety of baselines. HER \citep{her} is a model-free RL algorithm. SORB \citep{sorb} is a method that combines RL and graph search by using the entire replay buffer. Mapping State Space (MSS~\citealt{mss}) reduces the number of vertices by sub-sampling the replay buffer. \OurMethod, SORB, and MSS all use the same hindsight relabelling strategy proposed in HER. All of the domains are continuous control tasks, so we adopt DDPG \citep{ddpg} as the learning algorithm for the low-level actor. 

\subsection{Generalization to Longer Horizons}
\vspace{-2mm}
The PointMaze-Hard and AntMaze-Hard environments introduced in Figure \ref{fig:plan_visualize} are designed to test an agent's ability to generalize to longer horizons. While PointMaze and AntMaze have been previously used in \cite{duan2016benchmarking, mss, mega}, we make slight changes to those environments in order to increase their difficulty. We use a short, 200-timestep time horizon during training and a \(\rho_0\) that is uniform in the maze. At test time, we always initialize the agent on one end of the maze, and set the goal on the other end. The horizon of the test environment is 500 steps. Crucially, no prior knowledge on the shape of the maze is given to the agent. We also set a much stricter threshold for determining whether an agent has reached the goal. In Figure \ref{fig:test_curves}, we see \OurMethod is the only algorithm capable of solving AntMaze-Hard consistently. 

We observe an interesting trend where the success rates for some of other graph search methods crash and then slowly recover after making some initial progress. We postulate this occurs because methods that are based on using the entire replay or sub-sampling the replay for landmark selection will struggle as the buffer size increases. For instance, in the AntMaze-Hard environment, MSS and SORB use 400 and tens of thousands of landmarks respectively, whereas \OurMethod obtains a lean graph that only contain 50 learnable landmarks. The result suggests that \textit{learning latent landmarks} is significantly more sample efficient and stable than either directly using or sub-sampling the replay buffer to build the graph. The online planning algorithm in \OurMethod, which effectively leverages temporal abstraction to improve robustness, also contributes to the asymptotic success rate. As explained in Figure \ref{fig:plan_visualize}, \OurMethod successfully addresses the common failure modes of graph-based RL methods. The result convincingly shows that, at least on the navigation tasks considered, \OurMethod is most effective at taking advantage of the problem's inherent graph structure (without any prior knowledge of the map or environment configurations) and generalizing to longer horizons during test time. 

\subsection{Robotic Manipulation Tasks}
\vspace{-2mm}
We also benchmark challenging robotic manipulations tasks with a Fetch robot introduced in \cite{plappert2018multi, her}. Besides the PickAndPlace task, we also evaluate our method on two additional Fetch tasks involving a box on a table, as illustrated in Figure \ref{fig:place-task}. In Box-Distractor-PickAndPlace environment, the agent needs to perform the pick-and-place task with a box in the middle of the table serving as a distractor. The Place-Inside-Box environment aims to teach the agent to place an object with randomly initialized locations into the box and has a simple curriculum. During training, the goal distribution has 80\% regular pick-and-place goals, enabling the agent to first learn how to fetch in general. Meanwhile, only 20\% of the goals are inside the box, which is the harder part of the task. During testing, we evaluate the agent's ability to pick up the object from the table and place it inside the box. Our method achieves dominant performance in both learning speed and test-time generalization on those three robotic manipulation environments. We note that on those manipulation tasks considered, many prior planning methods \textit{hurt} the performance of the model-free agent. \OurMethod is the only method that is able to help the model-free agent learn faster and perform better on all three tasks. 


\subsection{Understanding Model Choices in \OurMethod}

We investigate \OurMethod's sensitivity to different design choices and hyper-parameters via a set of ablation studies. More specifically, we study how the following four factors affect the performance of \OurMethod: the choice of graph search algorithms, and edge weight cutoff threshold in graph search (a key hyper-parameter in the graph search module); the choice of online planning algorithms, and the number of latent landmarks being learned (a key hyper-parameter in the planning module). 

\begin{figure}
    \centering
    \includegraphics[width=0.35\textwidth]{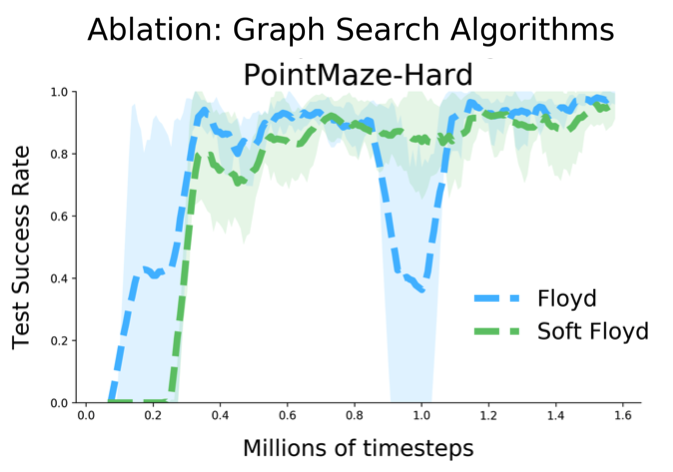}
    \vspace{-1em}
\end{figure}
\begin{figure}
    \centering
    \includegraphics[width=0.35\textwidth]{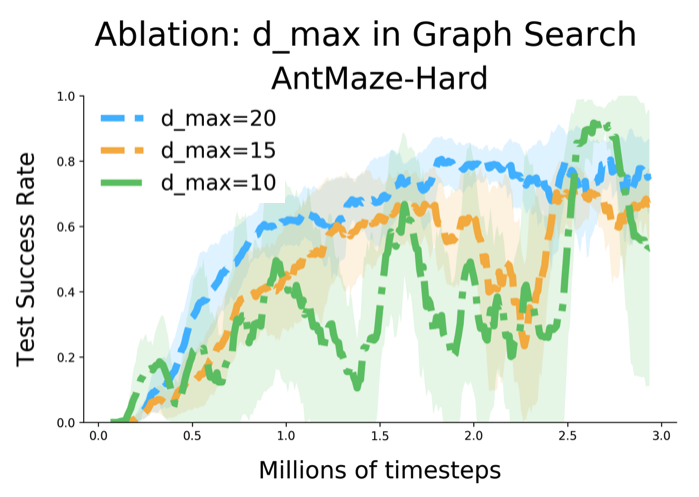}
    \caption{Ablation studies on the graph search module, including the choice of graph search algorithms and a key hyper-parameter in graph search: the edge weight cutoff threshold. }
    \label{fig:graph-ablation}
\end{figure}

While \OurMethod is agnostic to the graph search algorithm being used, we study the effect of two possible choices: Floyd algorithm and a soft version of Floyd (soft Floyd). As shown in Figure \ref{fig:graph-ablation}, the choice seems to have a relatively small effect on learning. During the early phase of experimentation, we find that having a \textit{soft} operation for relaxation in Floyd leads to better overall training stability. A hard version of relaxation helps the learning curve take off faster but suffers from greater instability during policy improvement. The likely reason is that neural distance estimates are not entirely accurate, and in the presence of occasional bad edges, using $\textit{softmax}$ rather than hard max improves robustness. We therefore use soft relaxation in \OurMethod. 

\begin{figure}
    \centering
    \includegraphics[width=0.34\textwidth]{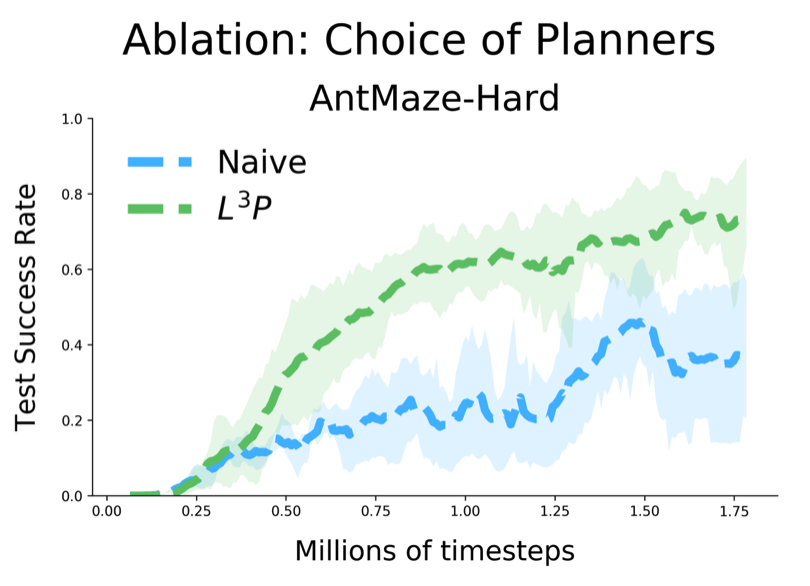}
    \vspace{-1em}
\end{figure}
\begin{figure}
    \centering
    \includegraphics[width=0.35\textwidth]{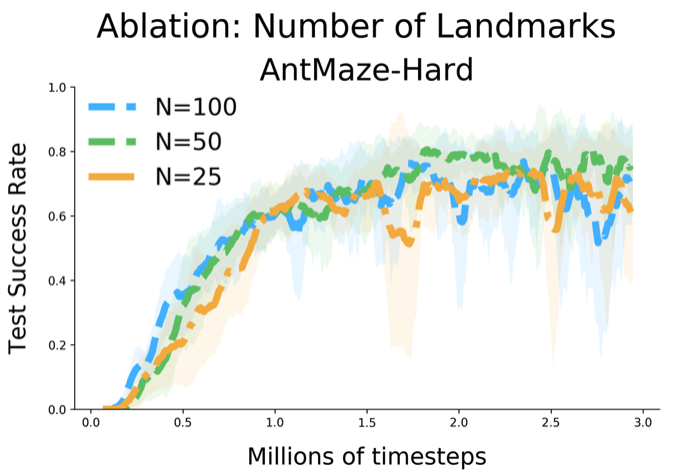}
    \caption{Ablation studies on the online planning module, including the choice of planners and a key hyper-parameter in graph-based planning: the number of nodes (landmarks).}
    \label{fig:ablation-planning}
\end{figure}

In the graph search module, a very sensitive hyper-parameter is the edge weight cutoff threshold, denoted as $d\_max$. This clipping threshold is commonly used in prior works such as \citep{sptm,sorb,mss,sgm}. It essentially means that if the weight of an edge is bigger than $d\_max$, then it is set to be infinity during the graph search process. The motivation for introducing this common hyper-parameter is two-fold. Firstly, we only trust distance estimates when they are \textit{local}, because value iterations are inherently local. Secondly, we want the next sub-goal to be relatively \textit{nearby} in terms of reachability. The $d\_max$ value determines the maximum (perceived) distance from the current state to next proposed subgoal. As shown in Figure \ref{fig:graph-ablation}, our current approach is still quite sensitive to this hyper-parameter; changes to $d\_max$ can have a considerable impact on learning. As this weakness is common to this class of approaches, we believe that further research is required to discover more principled ways of encouraging the search results to be local. 

For online planning, the \OurMethod planner introduced in Algorithm \autoref{online_planning} is essential to the success of \OurMethod. Our planning algorithm can take advantage of the temporal abstraction provided by the graph-structured world model. As previously shown in Figure \ref{fig:plan_visualize}, the design of \OurMethod planner avoids many common pitfalls. It does not re-plan at every step, but instead uses the reachability estimates to dynamically decide when to re-plan, striking a balance between adaptability and consistency in planning. This planner is also more tolerant of errors: it removes the immediate previous landmark when it re-plans, so that the agent will be less prone to getting stuck. In Figure \ref{fig:ablation-planning}, we compare the \OurMethod planner to a naive planner, which simply re-calculates the shortest path at every step. The result shows that our planning algorithm is crucial to the success of \OurMethod. 

An important hyper-parameter in graph-based planning is the number of landmarks being used. Intuitively, since \OurMethod is \textit{learning} the nodes on the graph, it should be robust to the changes in the number of nodes (landmarks) being learned. In Figure \ref{fig:ablation-planning}, we show that this is indeed the case: \OurMethod is robust to the number of latent landmarks. In contrast to prior methods, \OurMethod is able to \textit{learn} the nodes (landmarks) used for graph search from the agent's own experience. We vary this hyper-parameter in the challenging AntMaze-Hard environment, and we find that \OurMethod is robust against a variety of values. This is expected, because the landmarks in the latent space of \OurMethod will try to be equally scattered across the goal space according to the learned reachability metric. As the number of landmarks decreases, the learning procedure will automatically push the landmarks to be further away from one another. 

\section{Closing Remarks} 


In this work, we introduce a way of learning graph-structured world models that endow agents with the ability to do temporally extended reasoning. The algorithm,  \OurMethod, learns a set of latent landmarks scattered across the goal space to enable scalable planning. We demonstrate that \OurMethod achieves significantly better sample efficiency, higher asymptotic performance, and generalization to longer horizons on a range of challenging robotic navigation and manipulation tasks. Here we briefly discuss two promising future directions. First, how can an agent quickly generate a set of plausible landmarks in a previously unseen environment? A lot of progress has been made on the topics of meta reinforcement learning and learning to explore; can \OurMethod be combined with meta learning techniques for fast landmarks generation? Second, can we learn graph-structured world models from offline datasets? Batch RL is a more realistic setting for many RL applications, since online interaction can be expensive in the real world. Applying \OurMethod to offline datasets might require a notion of uncertainty in different parts of the graph. 


\section*{Acknowledgements}

We thank the anonymous reviewers for providing helpful comments on the paper. Resources used in preparing this research were provided, in part, by the Province of Ontario, the Government of Canada through CIFAR, and companies sponsoring the Vector Institute for Artificial Intelligence (\url{www.vectorinstitute.ai/partners}). 

\nocite{langley00}

\bibliography{example_paper}

\begin{thebibliography}{51}
\providecommand{\natexlab}[1]{#1}
\providecommand{\url}[1]{\texttt{#1}}
\expandafter\ifx\csname urlstyle\endcsname\relax
  \providecommand{\doi}[1]{doi: #1}\else
  \providecommand{\doi}{doi: \begingroup \urlstyle{rm}\Url}\fi

\bibitem[Andrychowicz et~al.(2017)Andrychowicz, Wolski, Ray, Schneider, Fong,
  Welinder, McGrew, Tobin, Abbeel, and Zaremba]{her}
Andrychowicz, M., Wolski, F., Ray, A., Schneider, J., Fong, R., Welinder, P.,
  McGrew, B., Tobin, J., Abbeel, O.~P., and Zaremba, W.
\newblock Hindsight experience replay.
\newblock In \emph{Advances in neural information processing systems}, pp.\
  5048--5058, 2017.

\bibitem[Arthur \& Vassilvitskii(2007)Arthur and
  Vassilvitskii]{kmeans_plus_plus}
Arthur, D. and Vassilvitskii, S.
\newblock k-means++: The advantages of careful seeding.
\newblock \emph{Proceedings of the eighteenth annual ACM-SIAM symposium on
  Discrete algorithms}, 2007.

\bibitem[Corneil et~al.(2018)Corneil, Gerstner, and Brea]{state_tabulation}
Corneil, D., Gerstner, W., and Brea, J.
\newblock Efficient model-based deep reinforcement learning with variational
  state tabulation.
\newblock \emph{arXiv preprint arXiv:1802.04325}, 2018.

\bibitem[Coulom(2006)]{mcts}
Coulom, R.
\newblock Efficient selectivity and backup operators in monte-carlo tree
  search.
\newblock In \emph{International conference on computers and games}, pp.\
  72--83. Springer, 2006.

\bibitem[Dayan \& Hinton(1993)Dayan and Hinton]{feudal_hinton}
Dayan, P. and Hinton, G.~E.
\newblock Feudal reinforcement learning.
\newblock In \emph{Advances in neural information processing systems}, pp.\
  271--278, 1993.

\bibitem[Dijkstra et~al.(1959)]{dijkstra1959note}
Dijkstra, E.~W. et~al.
\newblock A note on two problems in connexion with graphs.
\newblock \emph{Numerische mathematik}, 1\penalty0 (1):\penalty0 269--271,
  1959.

\bibitem[Doran \& Michie(1966)Doran and Michie]{doran1966experiments}
Doran, J.~E. and Michie, D.
\newblock Experiments with the graph traverser program.
\newblock \emph{Proceedings of the Royal Society of London. Series A.
  Mathematical and Physical Sciences}, 294\penalty0 (1437):\penalty0 235--259,
  1966.

\bibitem[Duan et~al.(2016)Duan, Chen, Houthooft, Schulman, and
  Abbeel]{duan2016benchmarking}
Duan, Y., Chen, X., Houthooft, R., Schulman, J., and Abbeel, P.
\newblock Benchmarking deep reinforcement learning for continuous control.
\newblock In \emph{International Conference on Machine Learning}, pp.\
  1329--1338, 2016.

\bibitem[Durrant-Whyte \& Bailey(2006)Durrant-Whyte and
  Bailey]{durrant2006simultaneous}
Durrant-Whyte, H. and Bailey, T.
\newblock Simultaneous localization and mapping: part i.
\newblock \emph{IEEE robotics \& automation magazine}, 13\penalty0
  (2):\penalty0 99--110, 2006.

\bibitem[Emmons et~al.(2020)Emmons, Jain, Laskin, Kurutach, Abbeel, and
  Pathak]{sgm}
Emmons, S., Jain, A., Laskin, M., Kurutach, T., Abbeel, P., and Pathak, D.
\newblock Sparse graphical memory for robust planning.
\newblock \emph{arXiv preprint arXiv:2003.06417}, 2020.

\bibitem[Eysenbach et~al.(2019)Eysenbach, Salakhutdinov, and Levine]{sorb}
Eysenbach, B., Salakhutdinov, R.~R., and Levine, S.
\newblock Search on the replay buffer: Bridging planning and reinforcement
  learning.
\newblock In \emph{Advances in Neural Information Processing Systems}, pp.\
  15246--15257, 2019.

\bibitem[Farquhar et~al.(2017)Farquhar, Rockt{\"a}schel, Igl, and
  Whiteson]{Farquhar2017treeQN}
Farquhar, G., Rockt{\"a}schel, T., Igl, M., and Whiteson, S.
\newblock Treeqn and atreec: Differentiable tree-structured models for deep
  reinforcement learning.
\newblock October 2017.
\newblock URL \url{http://arxiv.org/abs/1710.11417}.

\bibitem[Gillner \& Mallot(1998)Gillner and Mallot]{gillner1998navigation}
Gillner, S. and Mallot, H.~A.
\newblock Navigation and acquisition of spatial knowledge in a virtual maze.
\newblock \emph{Journal of cognitive neuroscience}, 10\penalty0 (4):\penalty0
  445--463, 1998.

\bibitem[Ha \& Schmidhuber(2018)Ha and Schmidhuber]{ha2018recurrent}
Ha, D. and Schmidhuber, J.
\newblock Recurrent world models facilitate policy evolution.
\newblock In \emph{Advances in Neural Information Processing Systems}, pp.\
  2450--2462, 2018.

\bibitem[Hafner et~al.(2019)Hafner, Lillicrap, Fischer, Villegas, Ha, Lee, and
  Davidson]{latent_dynamics}
Hafner, D., Lillicrap, T., Fischer, I., Villegas, R., Ha, D., Lee, H., and
  Davidson, J.
\newblock Learning latent dynamics for planning from pixels.
\newblock In \emph{International Conference on Machine Learning}, pp.\
  2555--2565. PMLR, 2019.

\bibitem[Hafner et~al.(2020)Hafner, Lillicrap, Norouzi, and
  Ba]{hafner2020mastering}
Hafner, D., Lillicrap, T., Norouzi, M., and Ba, J.
\newblock Mastering atari with discrete world models.
\newblock \emph{arXiv preprint arXiv:2010.02193}, 2020.

\bibitem[Hart et~al.(1968)Hart, Nilsson, and Raphael]{hart1968formal}
Hart, P.~E., Nilsson, N.~J., and Raphael, B.
\newblock A formal basis for the heuristic determination of minimum cost paths.
\newblock \emph{IEEE transactions on Systems Science and Cybernetics},
  4\penalty0 (2):\penalty0 100--107, 1968.

\bibitem[Huang et~al.(2019)Huang, Liu, and Su]{mss}
Huang, Z., Liu, F., and Su, H.
\newblock Mapping state space using landmarks for universal goal reaching.
\newblock In \emph{Advances in Neural Information Processing Systems}, pp.\
  1942--1952, 2019.

\bibitem[Janner et~al.(2019)Janner, Fu, Zhang, and Levine]{mbpo}
Janner, M., Fu, J., Zhang, M., and Levine, S.
\newblock When to trust your model: Model-based policy optimization.
\newblock In \emph{Advances in Neural Information Processing Systems}, pp.\
  12519--12530, 2019.

\bibitem[Jurgenson et~al.(2020)Jurgenson, Avner, Groshev, and
  Tamar]{subgoal-trees}
Jurgenson, T., Avner, O., Groshev, E., and Tamar, A.
\newblock Sub-goal trees--a framework for goal-based reinforcement learning.
\newblock \emph{arXiv preprint arXiv:2002.12361}, 2020.

\bibitem[Kingma \& Welling(2013)Kingma and Welling]{vae}
Kingma, D.~P. and Welling, M.
\newblock Auto-encoding variational bayes.
\newblock \emph{arXiv preprint arXiv:1312.6114}, 2013.

\bibitem[Kurutach et~al.(2018)Kurutach, Clavera, Duan, Tamar, and
  Abbeel]{me_trpo}
Kurutach, T., Clavera, I., Duan, Y., Tamar, A., and Abbeel, P.
\newblock Model-ensemble trust-region policy optimization.
\newblock \emph{arXiv preprint arXiv:1802.10592}, 2018.

\bibitem[Langley(2000)]{langley00}
Langley, P.
\newblock Crafting papers on machine learning.
\newblock In Langley, P. (ed.), \emph{Proceedings of the 17th International
  Conference on Machine Learning (ICML 2000)}, pp.\  1207--1216, Stanford, CA,
  2000. Morgan Kaufmann.

\bibitem[LaValle(1998)]{lavalle1998rapidly}
LaValle, S.~M.
\newblock Rapidly-exploring random trees: A new tool for path planning.
\newblock 1998.

\bibitem[Lee et~al.(2018)Lee, Parisotto, Chaplot, Xing, and
  Salakhutdinov]{Lee2018gppn}
Lee, L., Parisotto, E., Chaplot, D.~S., Xing, E., and Salakhutdinov, R.
\newblock Gated path planning networks.
\newblock June 2018.
\newblock URL \url{http://arxiv.org/abs/1806.06408}.

\bibitem[Lillicrap et~al.(2015)Lillicrap, Hunt, Pritzel, Heess, Erez, Tassa,
  Silver, and Wierstra]{ddpg}
Lillicrap, T.~P., Hunt, J.~J., Pritzel, A., Heess, N., Erez, T., Tassa, Y.,
  Silver, D., and Wierstra, D.
\newblock Continuous control with deep reinforcement learning.
\newblock \emph{arXiv preprint arXiv:1509.02971}, 2015.

\bibitem[Liu et~al.(2019)Liu, Kurutach, Tung, Abbeel, and Tamar]{Liu2019-mh}
Liu, K., Kurutach, T., Tung, C. K.-C., Abbeel, P., and Tamar, A.
\newblock Hallucinative topological memory for zero-shot visual planning.
\newblock \emph{ArXiv}, 2019.
\newblock URL \url{https://arxiv.org/abs/2002.12336}.

\bibitem[Luo et~al.(2018)Luo, Xu, Li, Tian, Darrell, and Ma]{slbo}
Luo, Y., Xu, H., Li, Y., Tian, Y., Darrell, T., and Ma, T.
\newblock Algorithmic framework for model-based deep reinforcement learning
  with theoretical guarantees.
\newblock \emph{arXiv preprint arXiv:1807.03858}, 2018.

\bibitem[Nachum et~al.(2018)Nachum, Gu, Lee, and Levine]{hiro}
Nachum, O., Gu, S.~S., Lee, H., and Levine, S.
\newblock Data-efficient hierarchical reinforcement learning.
\newblock In \emph{Advances in Neural Information Processing Systems}, pp.\
  3303--3313, 2018.

\bibitem[Nagabandi et~al.(2018)Nagabandi, Kahn, Fearing, and Levine]{mb_mf}
Nagabandi, A., Kahn, G., Fearing, R.~S., and Levine, S.
\newblock Neural network dynamics for model-based deep reinforcement learning
  with model-free fine-tuning.
\newblock In \emph{2018 IEEE International Conference on Robotics and
  Automation (ICRA)}, pp.\  7559--7566. IEEE, 2018.

\bibitem[Nair et~al.(2018)Nair, Pong, Dalal, Bahl, Lin, and Levine]{rig}
Nair, A.~V., Pong, V., Dalal, M., Bahl, S., Lin, S., and Levine, S.
\newblock Visual reinforcement learning with imagined goals.
\newblock In \emph{Advances in Neural Information Processing Systems}, pp.\
  9191--9200, 2018.

\bibitem[Nasiriany et~al.(2019)Nasiriany, Pong, Lin, and Levine]{leap}
Nasiriany, S., Pong, V., Lin, S., and Levine, S.
\newblock Planning with goal-conditioned policies.
\newblock In \emph{Advances in Neural Information Processing Systems}, pp.\
  14843--14854, 2019.

\bibitem[Pertsch et~al.(2020)Pertsch, Rybkin, Ebert, Finn, Jayaraman, and
  Levine]{long_plan}
Pertsch, K., Rybkin, O., Ebert, F., Finn, C., Jayaraman, D., and Levine, S.
\newblock Long-horizon visual planning with goal-conditioned hierarchical
  predictors.
\newblock \emph{arXiv preprint arXiv:2006.13205}, 2020.

\bibitem[Pitis et~al.(2020)Pitis, Chan, Zhao, Stadie, and Ba]{mega}
Pitis, S., Chan, H., Zhao, S., Stadie, B., and Ba, J.
\newblock Maximum entropy gain exploration for long horizon multi-goal
  reinforcement learning.
\newblock \emph{arXiv preprint arXiv:2007.02832}, 2020.

\bibitem[Plappert et~al.(2018)Plappert, Andrychowicz, Ray, McGrew, Baker,
  Powell, Schneider, Tobin, Chociej, Welinder, et~al.]{plappert2018multi}
Plappert, M., Andrychowicz, M., Ray, A., McGrew, B., Baker, B., Powell, G.,
  Schneider, J., Tobin, J., Chociej, M., Welinder, P., et~al.
\newblock Multi-goal reinforcement learning: Challenging robotics environments
  and request for research.
\newblock \emph{arXiv preprint arXiv:1802.09464}, 2018.

\bibitem[Pong et~al.(2018)Pong, Gu, Dalal, and Levine]{tdm}
Pong, V., Gu, S., Dalal, M., and Levine, S.
\newblock Temporal difference models: Model-free deep rl for model-based
  control.
\newblock \emph{arXiv preprint arXiv:1802.09081}, 2018.

\bibitem[Pong et~al.(2019)Pong, Dalal, Lin, Nair, Bahl, and Levine]{skew-fit}
Pong, V.~H., Dalal, M., Lin, S., Nair, A., Bahl, S., and Levine, S.
\newblock Skew-fit: State-covering self-supervised reinforcement learning.
\newblock \emph{arXiv preprint arXiv:1903.03698}, 2019.

\bibitem[Racani{\`e}re et~al.(2017)Racani{\`e}re, Weber, Reichert, Buesing,
  Guez, Jimenez~Rezende, Puigdom{\`e}nech~Badia, Vinyals, Heess, Li, Pascanu,
  Battaglia, Hassabis, Silver, and Wierstra]{Racaniere2017I2A}
Racani{\`e}re, S., Weber, T., Reichert, D., Buesing, L., Guez, A.,
  Jimenez~Rezende, D., Puigdom{\`e}nech~Badia, A., Vinyals, O., Heess, N., Li,
  Y., Pascanu, R., Battaglia, P., Hassabis, D., Silver, D., and Wierstra, D.
\newblock Imagination-augmented agents for deep reinforcement learning.
\newblock \emph{Neural Information Processing Systems}, 2017.

\bibitem[Savinov et~al.(2018{\natexlab{a}})Savinov, Dosovitskiy, and
  Koltun]{sptm}
Savinov, N., Dosovitskiy, A., and Koltun, V.
\newblock Semi-parametric topological memory for navigation.
\newblock \emph{arXiv preprint arXiv:1803.00653}, 2018{\natexlab{a}}.

\bibitem[Savinov et~al.(2018{\natexlab{b}})Savinov, Raichuk, Marinier, Vincent,
  Pollefeys, Lillicrap, and Gelly]{reachability}
Savinov, N., Raichuk, A., Marinier, R., Vincent, D., Pollefeys, M., Lillicrap,
  T., and Gelly, S.
\newblock Episodic curiosity through reachability.
\newblock \emph{arXiv preprint arXiv:1810.02274}, 2018{\natexlab{b}}.

\bibitem[Silver et~al.(2016{\natexlab{a}})Silver, Huang, Maddison, Guez, Sifre,
  Van Den~Driessche, Schrittwieser, Antonoglou, Panneershelvam, Lanctot,
  et~al.]{silver2016mastering}
Silver, D., Huang, A., Maddison, C.~J., Guez, A., Sifre, L., Van Den~Driessche,
  G., Schrittwieser, J., Antonoglou, I., Panneershelvam, V., Lanctot, M.,
  et~al.
\newblock Mastering the game of go with deep neural networks and tree search.
\newblock \emph{nature}, 529\penalty0 (7587):\penalty0 484--489,
  2016{\natexlab{a}}.

\bibitem[Silver et~al.(2016{\natexlab{b}})Silver, van Hasselt, Hessel, Schaul,
  Guez, Harley, Dulac-Arnold, Reichert, Rabinowitz, Barreto, and
  Degris]{Silver2016predictron}
Silver, D., van Hasselt, H., Hessel, M., Schaul, T., Guez, A., Harley, T.,
  Dulac-Arnold, G., Reichert, D., Rabinowitz, N., Barreto, A., and Degris, T.
\newblock The predictron: End-to-end learning and planning.
\newblock December 2016{\natexlab{b}}.
\newblock URL \url{http://arxiv.org/abs/1612.08810}.

\bibitem[Srinivas et~al.(2018)Srinivas, Jabri, Abbeel, Levine, and
  Finn]{Srinivas2018UPN}
Srinivas, A., Jabri, A., Abbeel, P., Levine, S., and Finn, C.
\newblock Universal planning networks.
\newblock April 2018.
\newblock URL \url{http://arxiv.org/abs/1804.00645}.

\bibitem[Sutton(1991)]{dyna}
Sutton, R.~S.
\newblock Dyna, an integrated architecture for learning, planning, and
  reacting.
\newblock \emph{ACM Sigart Bulletin}, 2\penalty0 (4):\penalty0 160--163, 1991.

\bibitem[Tamar et~al.(2016)Tamar, Wu, Thomas, Levine, and Abbeel]{tamar2016VIN}
Tamar, A., Wu, Y., Thomas, G., Levine, S., and Abbeel, P.
\newblock Value iteration networks.
\newblock February 2016.
\newblock URL \url{http://arxiv.org/abs/1602.02867}.

\bibitem[Vezhnevets et~al.(2017)Vezhnevets, Osindero, Schaul, Heess, Jaderberg,
  Silver, and Kavukcuoglu]{vezhnevets2017feudal}
Vezhnevets, A.~S., Osindero, S., Schaul, T., Heess, N., Jaderberg, M., Silver,
  D., and Kavukcuoglu, K.
\newblock Feudal networks for hierarchical reinforcement learning.
\newblock \emph{arXiv preprint arXiv:1703.01161}, 2017.

\bibitem[Wang \& Spelke(2002)Wang and Spelke]{wang2002human}
Wang, R.~F. and Spelke, E.~S.
\newblock Human spatial representation: Insights from animals.
\newblock \emph{Trends in cognitive sciences}, 6\penalty0 (9):\penalty0
  376--382, 2002.

\bibitem[Wang \& Ba(2019)Wang and Ba]{exploring_mbrl}
Wang, T. and Ba, J.
\newblock Exploring model-based planning with policy networks.
\newblock \emph{arXiv preprint arXiv:1906.08649}, 2019.

\bibitem[Wang et~al.(2008)Wang, Mulvaney, Sillitoe, and Swere]{wang2008robot}
Wang, Y., Mulvaney, D., Sillitoe, I., and Swere, E.
\newblock Robot navigation by waypoints.
\newblock \emph{Journal of Intelligent and Robotic Systems}, 52\penalty0
  (2):\penalty0 175--207, 2008.

\bibitem[Yang et~al.(2020)Yang, Zhang, Morcos, Pineau, Abbeel, and
  Calandra]{yang2020plan2vec}
Yang, G., Zhang, A., Morcos, A.~S., Pineau, J., Abbeel, P., and Calandra, R.
\newblock Plan2vec: Unsupervised representation learning by latent plans.
\newblock In \emph{Proceedings of The 2nd Annual Conference on Learning for
  Dynamics and Control}, volume 120 of \emph{Proceedings of Machine Learning
  Research}, pp.\  1--12, 2020.
\newblock arXiv:2005.03648.

\bibitem[Zhao et~al.(2019)Zhao, Sun, and Tresp]{mep}
Zhao, R., Sun, X., and Tresp, V.
\newblock Maximum entropy-regularized multi-goal reinforcement learning.
\newblock \emph{arXiv preprint arXiv:1905.08786}, 2019.

\end{thebibliography}
\bibliographystyle{icml2021}

\clearpage

\section*{Appendix A: Greedy Latent Sparsification}

\begin{figure}[h]
    \centering
    \includegraphics[width=0.50\textwidth]{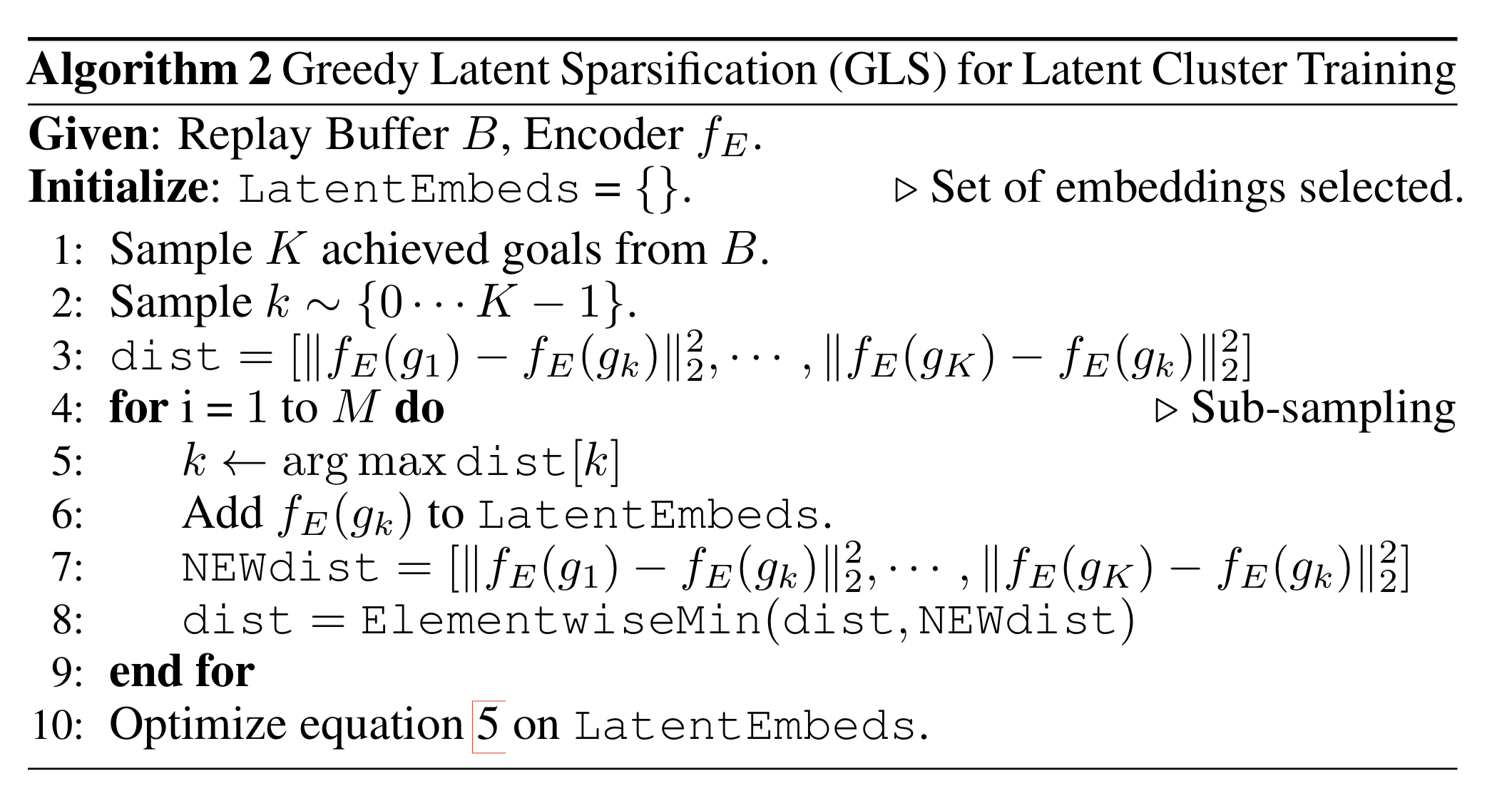}
    \vspace{-1em}
\end{figure}

The Greedy Latent Sparsification (GLS) algorithm sub-samples a large batch by sparsification. GLS first randomly selects a latent embedding from the batch, and then greedily chooses the next embedding that is furthest away from already selected embeddings. After collecting some \textit{warm-up trajectories} before planning starts (see Table 1 below) during training, we first use GLS to initialize the latent centroids, and then continue to use it to sample the batches used to train the latent clusters. GLS is strongly inspired by \cite{kmeans_plus_plus}, and this type of approach is known to improve clustering. 

\section*{Appendix B: Graph Search with Soft Relaxations} 

\label{appendix_sin}

In this paper, we employ a soft version of Floyd algorithm, which we find to empirically work well. Rather than simply using the $\min$ operation to do relaxation, the soft value iteration procedure uses a $soft\min$ operation when doing an update (note that, since we negated the distances to be negative in the weight matrix of the graph, the operations we use are actually max and softmax). The reason is that neural distances can be inconsistent and inaccurate at times, and using a soft operation makes the whole procedure more robust. More concretely, we repeat the following update on the weight matrix for $S$ steps with temperature $\beta$:
\begin{align}
    w_{i,j} &\leftarrow \sum_{k=1}^{N+1} \dfrac{\exp \dfrac{1}{\beta}(w_{i, k} + w_{k, j})}{ \sum_{k'=1}^{N+1} \exp \dfrac{1}{\beta}(w_{i, k'} + w_{k', j}) } \Big( w_{i, k} + w_{k, j} \Big) 
\end{align}
Following the practice in \cite{sorb, mss}, we do the following initialization to the distance matrix: for entries smaller than the negative of $d\_max$, we penalize the entry by adding $-\infty$ to it (in this paper, we use $-10^{6}$ as the $-\infty$ value). The essential idea is that we only trust a neural estimate when it is \textit{local}, and we rely on graph search to solve for \textit{global}, longer-horizon distances. The $-\infty$ penalty effectively masks out those entries with large negative values in the softmax operation above. If we replace softmax with a hard max, we recover the original update in Floyd algorithm; we can interpolate between a hard Floyd and a soft Floyd by tuning the temperature $\beta$. 

\section*{Appendix C: Overall Training Procedure} 

Here we provide an overall training procedure for \OurMethod in \textbf{Algorithm 3}. Given an environment \texttt{env} and a training goal distribution $p(g)$, we initialize a replay buffer $B$ and the following \textbf{trainble modules}: policy $\pi$, distance function $D$, value function $V$, encoder $f_{E}$ and decoder $f_{D}$, latent centroids $\{\textbf{c}_{1} \cdots \textbf{c}_{N}\}$. 

Every $K_{env}$ episodes of sampling, we take gradient steps for the above modules. The ratio between the number of environment steps and the number of gradient steps is a hyper-parameter. 

\begin{figure}[h]
    \centering
    \includegraphics[width=0.50\textwidth]{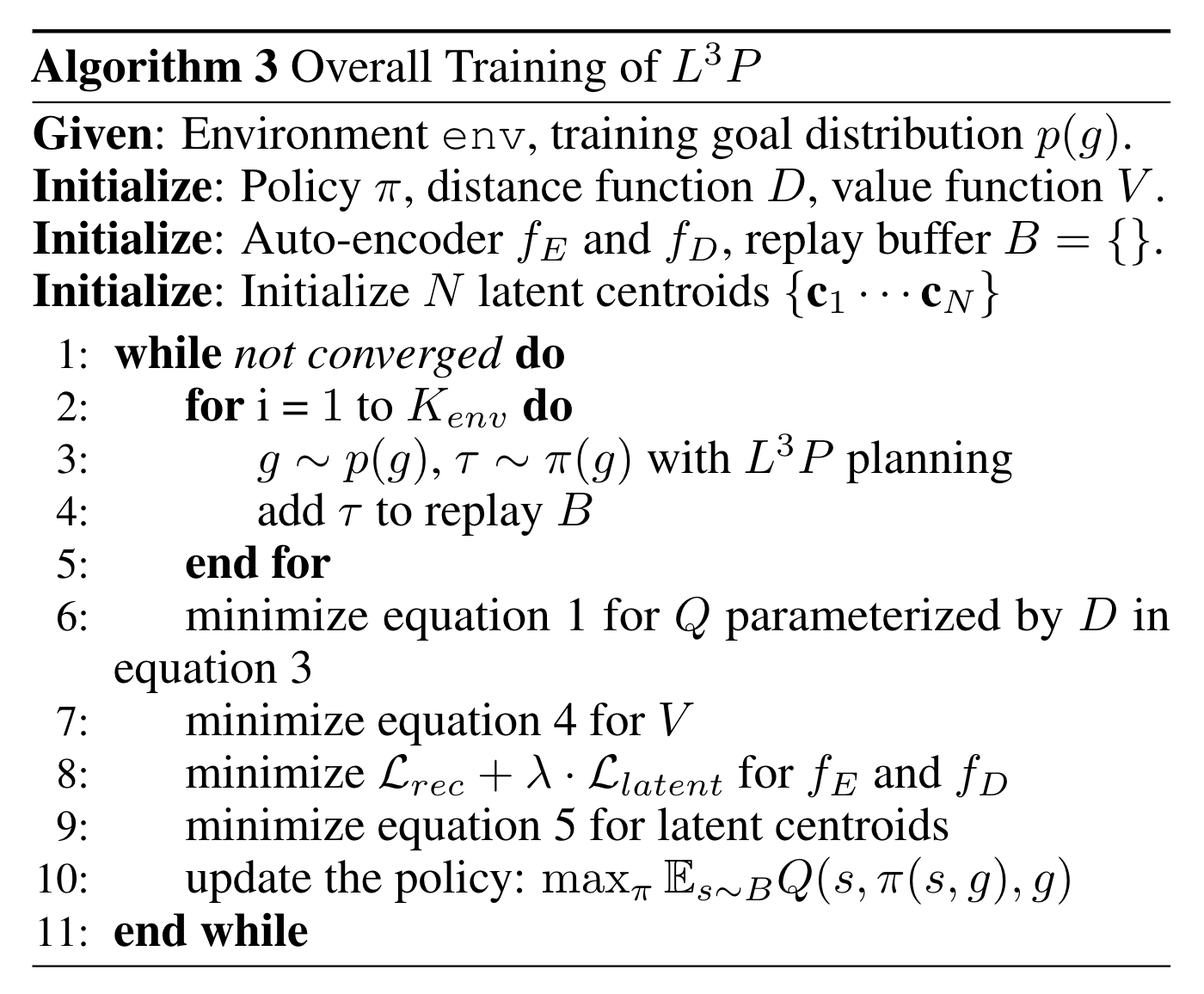}
    \vspace{-1em}
\end{figure}

\section*{Appendix D: Implementation Details}


\begin{itemize}
    \item We find that having a centralized replay for all parallel workers is significantly more sample efficient than having separate replays for each worker and simply averaging the gradients across workers. 
    \item For Ant-Maze environment, we do grad norm clipping by a value of $15.0$ for all networks. For Fetch tasks, we normalize the inputs by running means and standard deviations per input dimensions.
    \item Since \OurMethod is able to decompose a long-horizon goal into many short-horizon goals, we shorten the range of future steps where we do hindsight relabelling; as a result, the agent can focus its optimization effort on more immediate goals. This corresponds to the hyper-parameter: hindsight relabelling range. 
    \item During training, we collect $50\%$ of the data without the planning module, and the other $50\%$ of the data with planning. This corresponds to the hyper-parameter: probability of using search during train. 
    \item At train time, to encourage exploration during planning, we temporarily add a small number of random landmarks from GLS (\textbf{Algorithm 2}) to the existing latent landmarks. A new set of random landmarks is selected for each episode before graph search starts (\textbf{Algorithm 1}). This corresponds to the hyper-parameter: random landmarks added during train.
    \item We find that collecting a certain number of \textit{warm-up trajectories} for every worker before the planning procedure starts (during training) and before GLS (Algorithm 2) is used for initialization to help improve the planning results. This corresponds to the hyper-parameter: number of \textit{warm-up trajectories}.
\end{itemize}

\section*{Appendix E: Hyper-parameters}
\label{app:hypers}
The first table below lists the common hyper-parameters across all environments. The second table below lists the hyper-parameters that differ across the environments. 

\begin{figure}[h]
    \centering
    \includegraphics[width=0.50\textwidth]{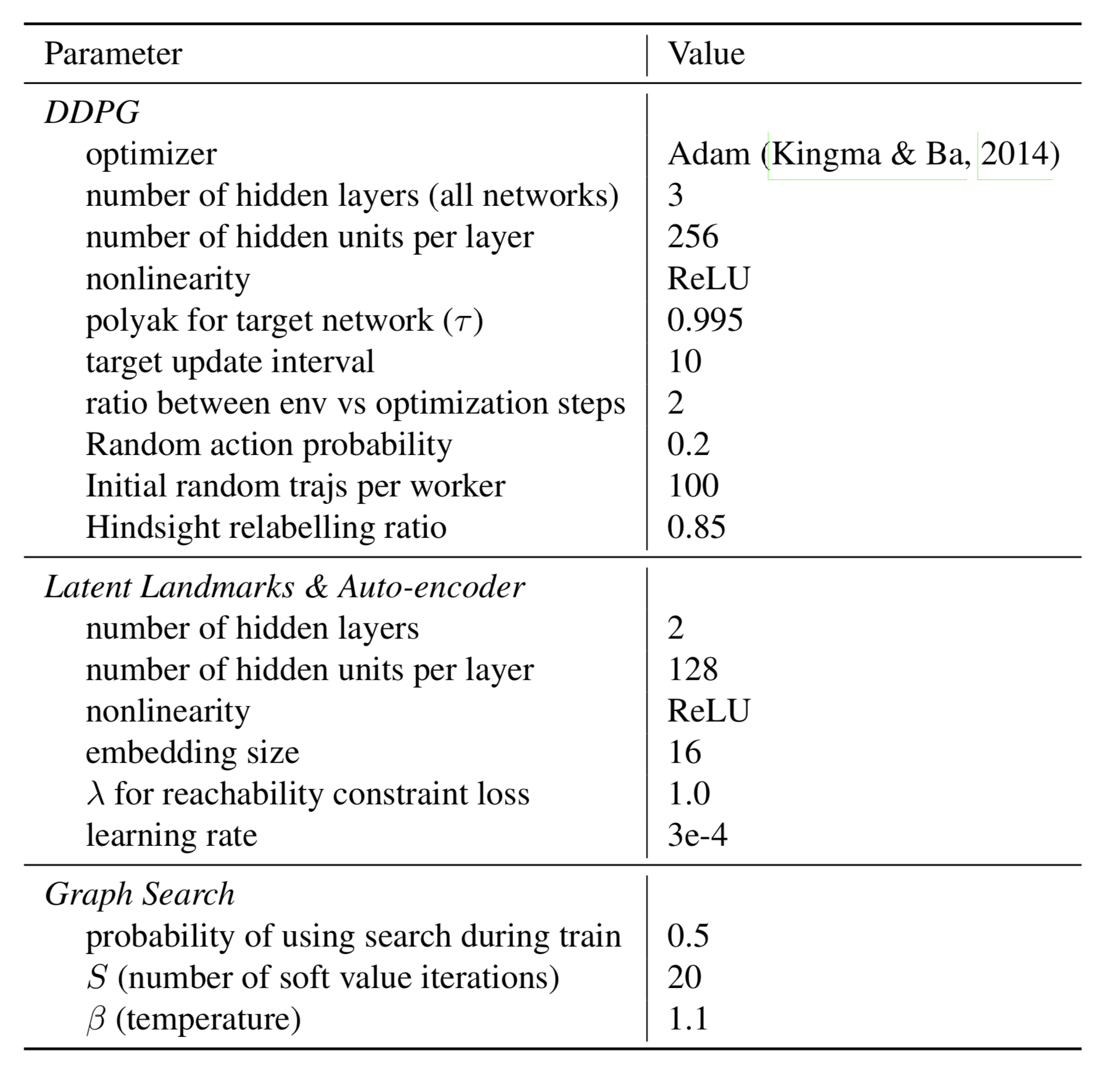}
    \vspace{-1em}
\end{figure}

\begin{figure}[h]
    \centering
    \includegraphics[width=0.50\textwidth]{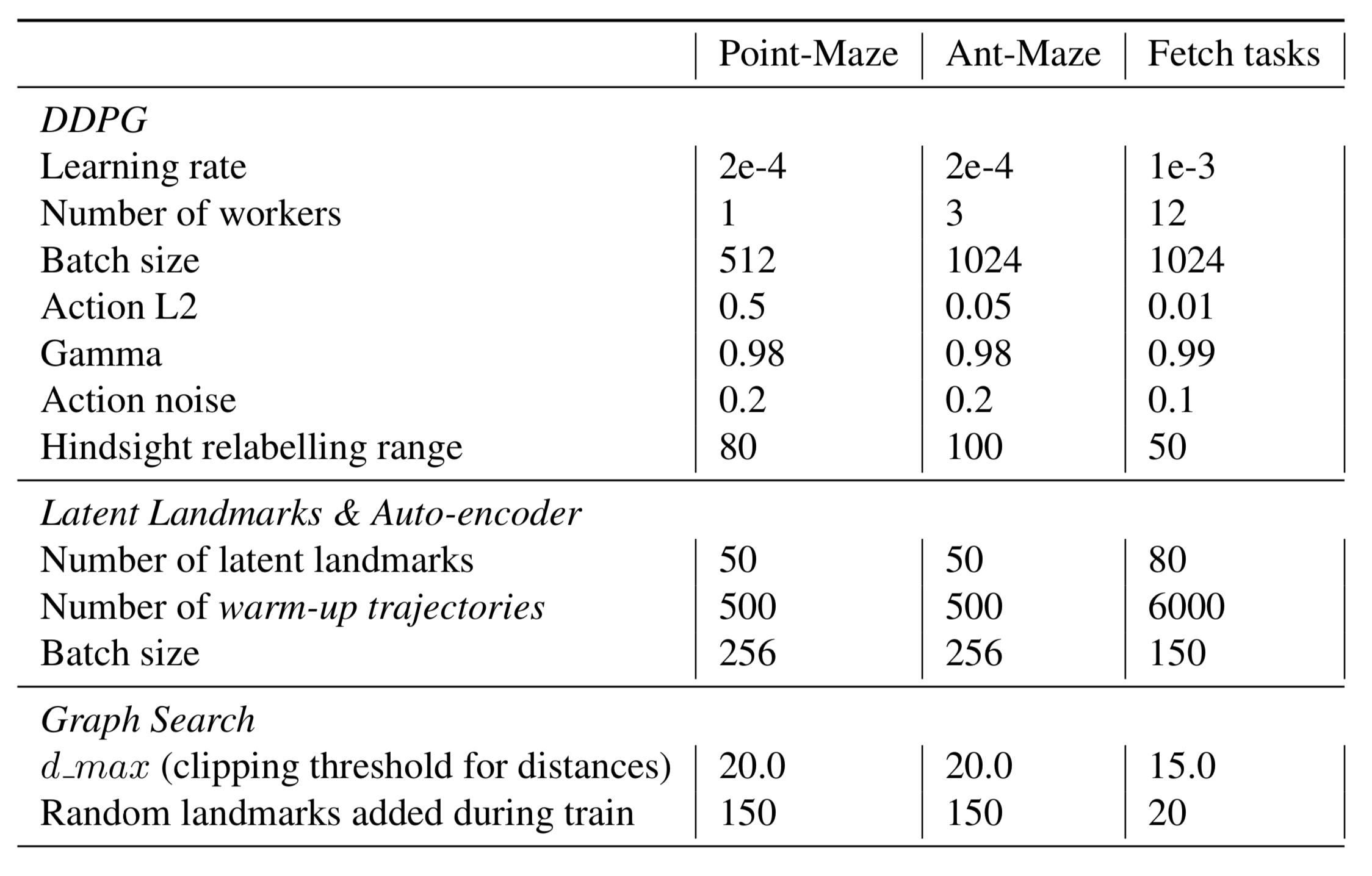}
    \vspace{-1em}
\end{figure}

\end{document}